\documentclass[11pt]{article}

\usepackage[preprint]{acl}

\usepackage{times}
\usepackage{latexsym}
\usepackage{multirow}
\usepackage[inline]{enumitem}
\usepackage{soul} 
\usepackage{booktabs}

\newcommand{\bo}[1]{\textcolor{black}{#1}}
\newcommand{\bocr}[1]{\textcolor{black}{#1}}

\usepackage[T1]{fontenc}

\usepackage[utf8]{inputenc}

\usepackage{microtype}

\usepackage{inconsolata}

\usepackage{graphicx}
\usepackage{url}
\usepackage{amsmath}
\usepackage[table]{xcolor}

\title{Argument Quality Assessment with Large Language Models: A Pairwise Bradley-Terry Approach}

\author{
  \textbf{Nicolás Benjamín Ocampo\textsuperscript{1}}, 
  \textbf{Agnes Paullate Nyiranziza\textsuperscript{2}},
  \textbf{Davide Ceolin\textsuperscript{1}}
\\
  \textsuperscript{1}HCDA, Centrum Wiskunde \& Informatica, Amsterdam, The Netherlands \\
  \textsuperscript{2}Vrije Universiteit Amsterdam, Amsterdam, The Netherlands
\\
  \small{
    \texttt{n.b.ocampo@cwi.nl, paullate.nziza@gmail.com, d.ceolin@cwi.nl}
  }
}

\begin{document}
\maketitle
\begin{abstract}
Large Language Models (LLMs) have demonstrated remarkable capabilities in tasks related to reasoning and judgment. However, assessing the quality of arguments requires a rigorous evaluation. We investigate the extent to which LLMs can effectively perform this task.
\bo{We tested 12 open-weight LLMs of different sizes and families under zero-shot, few-shot, and chain-of-thought 
to approximate human pairwise comparisons of argument quality across three dimensions---logical, rhetorical, and dialectic---
and used these comparisons in a Bradley-Terry model to infer latent strength scores and derive a ranking of arguments.}
\bo{
Our insights show that LLMs have promising but moderate correlation with human 
judgment, with Llama-70B obtaining the strongest alignment, reaching moderate Cohen’s $\kappa$ = 0.493 and moderate correlations with Bradley–Terry scores derived from these annotations (Kendall, Pearson, and Spearman: 0.327–0.477). Other LLMs exhibit weak, moderate, or high alignment with Llama-70B while achieving comparable results against human judgment, suggesting partial but complementary understanding of underlying quality dimensions despite differences in model size and family. Moreover, LLM predictions are stable across trial runs, with fewer than 7.75\% of cases yielding different labels. Remaining variability is handled via majority voting and few-shot prompting for large-size models.
}
\end{abstract}

\section{Introduction}\label{s:intro}

Argumentation is fundamental to reasoning in various fields \cite{chalaguine2017assessing}: scientists publish discoveries together with supporting evidence, lawyers structure legal arguments to solve disputes, and political debaters rely on argumentation to gain the approval of the public. Early computational work focused on mining and evaluating arguments in well-structured texts such as essays \cite{wachsmuth2016using} until more research shifted toward assessing argument quality~\cite{habernal-gurevych-2016-argument, wachsmuth2017computational,toledo-etal-2019-automatic, vecchi-etal-2021-towards, joshi-etal-2023-arganalysis35k, wachsmuth2024argument}. A particularly effective approach within the field has been the use of pairwise comparisons~\cite{gienapp-etal-2020-efficient, habernal-gurevych-2016-argument, toledo-etal-2019-automatic, gretzetal}, in which the annotators judge which of the two arguments is stronger based on specific criteria. These judgments can then be aggregated using statistical models, such as PageRank~\cite{habernal-gurevych-2016-argument}, Gaussian Process~\cite{10.1162/tacl_a_00026}, and Bradley-Terry~\cite{gienapp-etal-2020-efficient}, to produce quality scores and obtain a ranking of evaluated arguments. 
This design reduces annotation cost and variance while providing reliable multidimensional quality scores for benchmarking automated evaluation methods.
As online platforms grow, the ability to identify high-quality arguments becomes increasingly important, especially for effective persuasion, decision-making, and participation in meaningful discourse~\cite{wachsmuth2024argument}. However, manually evaluating argument quality at scale remains a major bottleneck due to cost and scalability.

Large Language Models (LLMs) offer a promising avenue for scaling argument quality assessment. LLMs demonstrated impressive capabilities in language understanding tasks, and a growing line of work explores the use of LLMs as annotators~\cite{mirzakhmedova2024large}—an example of LLM-as-a-judge—where the LLM is asked to evaluate text quality~\cite{GU2026101253}. This strategy could enable large-scale assessments with minimal cost and time. However, whether LLMs can reliably replicate human-level judgments is an open question. Existing research offers mixed results: some studies show that \bo{LLMs}, 
such as GPT-3 and PaLM-2 can effectively assess certain quality aspects~\cite{mirzakhmedova2024large}, while others found that GPT-3.5 underperformed specialized supervised models on tasks that involve ranking of arguments~\cite{wang-etal-2023-contextual}. 
These contrasting findings motivate a deeper investigation into LLM-based argument quality evaluation. This study attempts to replicate the argument quality rankings from the \bo{Webis-ArgQuality-20 dataset \cite{gienapp-etal-2020-efficient}} that were obtained by modeling, via Bradley-Terry, pairwise comparisons annotations done by human crowdsourcers. Given that argument quality is multidimensional~\cite{wachsmuth2017computational}, this study will also examine the abilities of the LLMs in assessing the different quality dimensions.
We aim to address the following research question (RQ):

\emph{RQ: How well can LLM pairwise comparisons reproduce human rankings of argument quality?}

We break it down through the following sub-research questions (SRQ):
\begin{enumerate}[label=\arabic*),itemjoin={\quad}]
    \item \textbf{SRQ1}: \emph{How do LLMs of different sizes and families, when using various prompting strategies, compare in their ability to evaluate arguments on 
    quality dimensions 
    compared against human judgment?}
    
    \item \textbf{SRQ2}: \emph{To what extent do LLMs of varying sizes and model families exhibit similar patterns of argument evaluation?}
    \item \textbf{SRQ3}: \emph{How consistent are LLM-based evaluations when repeated over multiple trials?}
\end{enumerate}

\bo{Our contribution is an extensive evaluation of 12 open-weight LLMs from five model families (7B--104B parameters) under zero-shot, few-shot, and chain-of-thought 
for pairwise argument assessment across logical, rhetorical, and dialectical dimensions. Using LLM pairwise judgments, we derive Bradley-Terry rankings, being, to the best of our knowledge, the first application of Bradley-Terry ranking to LLM-based comparative judgments in argument mining. 
Leveraging this contribution, we derive three important insights: (i) Our results show that LLM-derived BT rankings achieve moderate correlation with human judgments (up to 0.327–0.477 across Kendall, Pearson, and Spearman), with moderate human-level pairwise agreement (up to Cohen’s $\kappa$=0.493), highlighting both the promise and current limitations of LLMs as evaluators of argument quality. (ii) Other LLMs show weak-to-high alignment with the best performing model, Llama-70B, while matching human performance, indicating complementary understanding of quality dimensions across models. (iii) LLM predictions are stable, with label differences in at most 7.75\% across three runs. Few-shot prompting further improves stability for larger models, whereas smaller models are inherently more stable but less accurate \footnote{Code, and LLM generations and rankings: \url{https://github.com/benjaminocampo/arg_quality_with_llms}.}.}

\section{Related Work}\label{s:related}


Argument quality assessment is relevant, as people tend to seek the strongest arguments to form opinions, write persuasive texts, foster agreement, and improve mutual understanding when presenting ideas to an audience~\cite{wachsmuth2024argument}. To address this need, several frameworks conceptualize argument quality as a multidimensional construct. For example, the taxonomy proposed by \citet{wachsmuth2017computational} distinguishes logical, rhetorical, and dialectical dimensions of argument quality, providing a strong theoretical foundation for computational assessment. Complementing this perspective,~\citet{vecchi-etal-2021-towards, https://doi.org/10.1002/poi3.95} emphasize deliberative norms that focus on the collaborative and constructive aspects of discourse, ~\citet{habernal-gurevych-2016-argument, toledo-etal-2019-automatic} focus on argument convincingness, and~\citet{joshi-etal-2023-arganalysis35k} operationalize argument quality through a recommendation criterion, namely whether an argument would be suitable for use, as-is, in a supporting or opposing speech.

Using these theoretical frameworks, several datasets for argument quality assessment have been proposed that vary in number of topics and arguments, ranking strategies, evaluation dimensions, research communities, number of annotators, and annotation methods~\cite{romberg-etal-2025-towards}. Two main approaches have been adopted for evaluation: Likert or continuous scale judgments of individual arguments \cite{swanson-etal-2015-argument, wachsmuth2017computational, 10.1145/3331184.3331327, lauscher-etal-2020-rhetoric} and, pairwise comparison setups in which annotators select the better of two arguments with respect to a given dimension \cite{habernal-gurevych-2016-argument, toledo-etal-2019-automatic, gretzetal, gienapp-etal-2020-efficient}. Pairwise comparisons avoid the problems of absolute ratings by not requiring assessors to assign fixed labels that are often treated as interval-scale, and instead asking them to simply choose between two items, which is easier, more consistent, and produces more reliable data~\cite{gienapp-etal-2020-efficient}. Subsequent scores are later obtained using a ranking function such as PageRank~\cite{Page1999ThePC} (or variants of it~\cite{habernal-gurevych-2016-argument}), Gaussian Process~\cite{10.1162/tacl_a_00026}, or Bradley-Terry~\cite{hunter2004mm, 10.1145/2433396.2433420}.

As pairwise comparisons scale rapidly, recent studies have explored using LLMs to assess argument quality. \citet{mirzakhmedova2024large, wang-etal-2023-contextual, deshpande-etal-2024-contextualizing} investigated agreement between LLMs and human annotators using models like GPT-3, GPT-3.5, and PaLM-2. 
\citet{rescala-etal-2024-language} examined LLMs' ability to assess argument convincingness, \citet{elaraby-etal-2024-persuasiveness} studied the role of rationales in pairwise quality judgments, \citet{jin-etal-2025-multi} proposed a multi-persona framework based on Llama models, and \citet{de-la-iglesia-etal-2025-ranking} used LLM-based ranking with medical arguments. Still, these works evaluate only a small set of LLMs, rely on Likert-scale, or focus on convincingness rather than multidimensional argument quality. In contrast, we systematically compare multiple LLM families and sizes across prompts using the broader theoretical framework of \citet{wachsmuth2017computational}, operationalized through the Webis-ArgQuality-20 dataset. We assess argument quality via pairwise comparisons and aggregate judgments with the Bradley-Terry model, a well-established and computationally efficient approach \cite{hunter2004mm, 10.1145/2433396.2433420}.

\section{Methodology}\label{s:background}



The overall methodology consists of performing comparisons of argument pairs using \bo{zero-shot, few-shot, and chain-of-thought}. 
Arguments are aggregated per topic (only topic-related arguments are compared), and the comparison is performed across three dimensions of argument quality.

\subsection{Argument Quality Dimensions}
\label{sec:dimensions}
Theoretical work in argumentation distinguishes three dimensions: logical, rhetorical, and dialectical \cite{blair2011groundwork, wachsmuth2017computational}. 
\textbf{Logical quality}, or \textit{cogency} checks whether premises  are relevant and sufficient to conclude. \citet{wachsmuth2017theoryvspractice} defines logical quality as: \textit{local acceptability} (are the premises worthy of acceptance?), \textit{local relevance} (do the premises support or attack the conclusion?), and \textit{local sufficiency} (are the premises enough to justify the conclusion?). 
\textbf{Rhetorical quality} or \textit{effectiveness} measures whether the argument can persuade a target audience. Inspired by Aristotle's rhetorical appeals, \citet{wachsmuth2017computational} further decomposes this dimension into: \textit{credibility} (establishing the speaker's authority), \textit{emotional appeal} (engaging the audience), \textit{clarity} (using correct and unambiguous language), \textit{appropriateness} (aligning the tone and language with the topic), and \textit{arrangement} (structuring the argument). 
\textbf{Dialectic quality}, or \textit{reasonableness}  
measures the ability of the argument to help resolve a disagreement. It is further decomposed into \textit{acceptability} (the audience's acceptance of the argument),  \textit{relevance} (the ability of the argument to advance the discussion), and \textit{ sufficiency} (rebuttal to counter arguments) \cite{wachsmuth2017theoryvspractice}. 

\subsection{Pairwise Comparison via Bradley-Terry}
Pairwise comparisons are a widely used method in statistical analysis. 
In argument quality assessment, this approach allows annotators to reduce cognitive load and annotation time by focusing on comparing two arguments at a time, instead of ranking a whole set. 
\bocr{Once we obtain these comparisons, we use the Bradley-Terry model to derive continuous quality scores. Given a pair of arguments $d_i$  and $d_j$, BT estimates the probability that $d_i$ is preferred to $d_j$ ($d_i \succ d_j$) through the formula~\cite{hunter2004mm}:}
\begin{equation*}
P(d_i \succ d_j) = \frac{\gamma_i}{\gamma_i +\gamma_j}
\end{equation*}

\bocr{If ties are allowed in pairwise comparisons, the Bradley--Terry model can be extended in several ways, including the formulations proposed by~\citet{Rao01031967, gienapp-etal-2020-efficient}. In our study, we instead adopted a simpler approximation, treating each tie as two reciprocal comparisons, one in each direction.} 



\section{Dataset}
We used the Webis-ArgQuality-20 corpus, an argument quality dataset introduced by~\citet{gienapp-etal-2020-efficient}, 
comprising 1271 arguments across 20 debate topics 
which were subjected to 13952 pairwise comparisons obtained via cyclic group sampling. 
\bocr{Crowdsourcers annotated, for each topic and quality dimension (Logical, Rhetorical, or Dialectical Quality), which argument in each pair was better with respect to that specific quality dimension. These ground-truth labels were subsequently used to rank argument quality across dimensions using a Bradley-Terry model \bocr{adjusted for ties}. Only one crowdsourcer per pair was used, so inter-annotator agreement for the comparisons in the dataset is not reported. Instead, to assess annotation quality, \citet{gienapp-etal-2020-efficient} relied on the Amazon Mechanical Turk HIT approval rate, requiring each annotator to have an approval rate of at least 95\%. This shifts the quality control mechanism to the annotator level.} We used the entire set of pairwise comparisons provided in the dataset in our experiments.

\section{Experimental setup}\label{s:experimental_setup}

\begin{figure}[!t]
  \centering
  \includegraphics[width=0.46\textwidth, trim=6.1cm 5.6cm 7.7cm 5.5cm,
  clip]{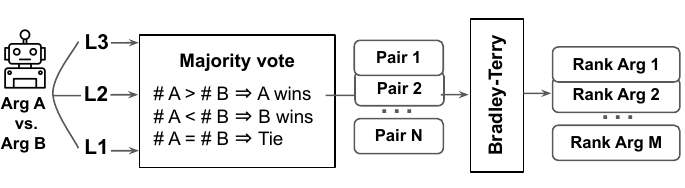}
  \caption{Pipeline overview of the ranking methodology. LLM pairwise comparison labels (L1–L3) are first resolved by majority vote. All the resulting pairwise labels are then aggregated with the Bradley–Terry model to produce a ranking of all arguments.}
  \label{fig:pipeline}
\end{figure}

\begin{table*}[t]
\centering
\scriptsize
\resizebox{!}{5.75cm}{\
\begin{tabular}{llccccccc}
\toprule
Model & Prompt & Pearson ($r$) & Spearman ($\rho$) & Kendall & MAE & RMSE & $\kappa$ IAA & \% Agreement \\
\midrule
\multirow[t]{3}{*}{Mistral-7B} & zero & 0.232 $\pm$ 0.021 & 0.232 $\pm$ 0.019 & 0.156 $\pm$ 0.012 & 0.973 $\pm$ 0.003 & 1.239 $\pm$ 0.017 & 0.162 $\pm$ 0.060 & 26.88 $\pm$ 5.16 \\
                               & few & 0.082 $\pm$ 0.183 & 0.082 $\pm$ 0.181 & 0.055 $\pm$ 0.123 & 1.046 $\pm$ 0.112 & 1.350 $\pm$ 0.138 & 0.332 $\pm$ 0.133 & 39.49 $\pm$ 9.85 \\
                               & cot & 0.221 $\pm$ 0.034 & 0.213 $\pm$ 0.033 & 0.143 $\pm$ 0.022 & 0.967 $\pm$ 0.021 & 1.248 $\pm$ 0.028 & 0.095 $\pm$ 0.051 & 22.39 $\pm$ 4.47 \\

\multirow[t]{3}{*}{Llama-8B} & zero & 0.258 $\pm$ 0.019 & 0.263 $\pm$ 0.018 & 0.176 $\pm$ 0.011 & 0.968 $\pm$ 0.010 & 1.218 $\pm$ 0.015 & 0.373 $\pm$ 0.016 & 42.68 $\pm$ 1.33 \\
                               & few & 0.155 $\pm$ 0.035 & 0.155 $\pm$ 0.024 & 0.104 $\pm$ 0.016 & 1.020 $\pm$ 0.009 & 1.300 $\pm$ 0.027 & 0.424 $\pm$ 0.006 & 46.62 $\pm$ 0.67 \\
                               & cot & 0.057 $\pm$ 0.051 & 0.056 $\pm$ 0.047 & 0.037 $\pm$ 0.032 & 1.080 $\pm$ 0.022 & 1.373 $\pm$ 0.037 & 0.303 $\pm$ 0.012 & 37.01 $\pm$ 1.01 \\

\multirow[t]{3}{*}{Olmo2-7B} & zero & -0.034 $\pm$ 0.044 & -0.032 $\pm$ 0.031 & -0.021 $\pm$ 0.021 & 1.121 $\pm$ 0.029 & 1.438 $\pm$ 0.031 & 0.331 $\pm$ 0.017 & 39.12 $\pm$ 1.36 \\
                               & few & -0.087 $\pm$ 0.032 & -0.103 $\pm$ 0.015 & -0.069 $\pm$ 0.010 & 1.149 $\pm$ 0.016 & 1.474 $\pm$ 0.022 & 0.401 $\pm$ 0.007 & 44.51 $\pm$ 0.74 \\
                                & cot & 0.002 $\pm$ 0.060 & -0.003 $\pm$ 0.084 & -0.002 $\pm$ 0.056 & 1.105 $\pm$ 0.048 & 1.412 $\pm$ 0.042 & 0.356 $\pm$ 0.031 & 41.07 $\pm$ 2.44 \\

\multirow[t]{3}{*}{Qwen2.5-7B} & zero & 0.322 $\pm$ 0.031 & 0.326 $\pm$ 0.033 & 0.221 $\pm$ 0.023 & 0.918 $\pm$ 0.018 & 1.164 $\pm$ 0.026 & 0.431 $\pm$ 0.010 & 47.51 $\pm$ 0.87 \\
                               & few & 0.354 $\pm$ 0.042 & 0.359 $\pm$ 0.032 & 0.244 $\pm$ 0.021 & 0.882 $\pm$ 0.022 & 1.137 $\pm$ 0.038 & 0.456 $\pm$ 0.004 & 49.57 $\pm$ 0.34 \\
                               & cot & 0.188 $\pm$ 0.155 & 0.201 $\pm$ 0.158 & 0.135 $\pm$ 0.108 & 1.007 $\pm$ 0.103 & 1.270 $\pm$ 0.123 & 0.399 $\pm$ 0.035 & 44.87 $\pm$ 2.96 \\

\multirow[t]{3}{*}{Command-R-7B} & zero & 0.370 $\pm$ 0.067 & 0.363 $\pm$ 0.068 & 0.247 $\pm$ 0.048 & 0.887 $\pm$ 0.051 & 1.122 $\pm$ 0.059 & 0.417 $\pm$ 0.007 & 46.34 $\pm$ 0.62 \\
                               & few & 0.301 $\pm$ 0.146 & 0.302 $\pm$ 0.145 & 0.205 $\pm$ 0.099 & 0.924 $\pm$ 0.097 & 1.178 $\pm$ 0.121 & 0.376 $\pm$ 0.011 & 43.04 $\pm$ 1.01 \\
                               & cot & 0.345 $\pm$ 0.031 & 0.344 $\pm$ 0.043 & 0.233 $\pm$ 0.030 & 0.903 $\pm$ 0.029 & 1.144 $\pm$ 0.027 & 0.340 $\pm$ 0.012 & 40.25 $\pm$ 0.99 \\
\midrule
\multirow[t]{3}{*}{Mixtral-8x7B} & zero & 0.174 $\pm$ 0.049 & 0.172 $\pm$ 0.047 & 0.115 $\pm$ 0.032 & 1.011 $\pm$ 0.021 & 1.285 $\pm$ 0.038 & 0.339 $\pm$ 0.033 & 40.09 $\pm$ 2.56 \\
                               & few & 0.233 $\pm$ 0.170 & 0.246 $\pm$ 0.150 & 0.166 $\pm$ 0.103 & 0.951 $\pm$ 0.094 & 1.233 $\pm$ 0.137 & 0.392 $\pm$ 0.063 & 44.22 $\pm$ 4.88 \\
                               & cot & 0.093 $\pm$ 0.119 & 0.084 $\pm$ 0.107 & 0.057 $\pm$ 0.073 & 1.047 $\pm$ 0.077 & 1.345 $\pm$ 0.088 & 0.277 $\pm$ 0.060 & 35.69 $\pm$ 3.80 \\

\multirow[t]{3}{*}{Mistral-22B} & zero & -0.086 $\pm$ 0.035 & -0.094 $\pm$ 0.016 & -0.063 $\pm$ 0.011 & 1.142 $\pm$ 0.009 & 1.473 $\pm$ 0.024 & 0.381 $\pm$ 0.003 & 42.90 $\pm$ 0.20 \\
                               & few & -0.047 $\pm$ 0.019 & -0.053 $\pm$ 0.012 & -0.035 $\pm$ 0.008 & 1.128 $\pm$ 0.009 & 1.447 $\pm$ 0.013 & 0.378 $\pm$ 0.007 & 42.65 $\pm$ 0.47 \\
                               & cot & -0.072 $\pm$ 0.060 & -0.084 $\pm$ 0.045 & -0.056 $\pm$ 0.030 & 1.141 $\pm$ 0.031 & 1.464 $\pm$ 0.041 & 0.370 $\pm$ 0.013 & 42.14 $\pm$ 0.94 \\

\multirow[t]{3}{*}{Olmo2-32B} & zero & 0.402 $\pm$ 0.018 & 0.377 $\pm$ 0.024 & 0.257 $\pm$ 0.017 & 0.860 $\pm$ 0.010 & 1.094 $\pm$ 0.017 & 0.366 $\pm$ 0.036 & 42.41 $\pm$ 2.87 \\
                               & few & 0.296 $\pm$ 0.081 & 0.295 $\pm$ 0.083 & 0.199 $\pm$ 0.057 & 0.937 $\pm$ 0.049 & 1.185 $\pm$ 0.068 & 0.392 $\pm$ 0.023 & 44.25 $\pm$ 1.86 \\
                               & cot & 0.398 $\pm$ 0.026 & 0.380 $\pm$ 0.028 & 0.258 $\pm$ 0.020 & 0.869 $\pm$ 0.020 & 1.097 $\pm$ 0.023 & 0.402 $\pm$ 0.028 & 45.29 $\pm$ 2.30 \\

\multirow[t]{3}{*}{Mixtral-8x22B} & zero & 0.237 $\pm$ 0.035 & 0.227 $\pm$ 0.023 & 0.152 $\pm$ 0.016 & 0.978 $\pm$ 0.022 & 1.235 $\pm$ 0.029 & 0.311 $\pm$ 0.055 & 37.95 $\pm$ 4.42 \\
                               & few & 0.290 $\pm$ 0.091 & 0.273 $\pm$ 0.083 & 0.185 $\pm$ 0.058 & 0.938 $\pm$ 0.068 & 1.190 $\pm$ 0.076 & 0.412 $\pm$ 0.016 & 45.95 $\pm$ 1.39 \\
                               & cot & 0.218 $\pm$ 0.029 & 0.217 $\pm$ 0.011 & 0.145 $\pm$ 0.008 & 1.000 $\pm$ 0.018 & 1.250 $\pm$ 0.024 & 0.371 $\pm$ 0.032 & 42.66 $\pm$ 2.28 \\
\midrule
\multirow[t]{3}{*}{Llama-70B} & zero & 0.442 $\pm$ 0.005 & 0.441 $\pm$ 0.012 & 0.301 $\pm$ 0.010 & 0.836 $\pm$ 0.002 & 1.056 $\pm$ 0.004 & \underline{0.487 $\pm$ 0.008} & 52.33 $\pm$ 0.77 \\
                               & few & \textbf{0.473 $\pm$ 0.004} & \textbf{0.477 $\pm$ 0.012} & \textbf{0.327 $\pm$ 0.008} & \textbf{0.811 $\pm$ 0.006} & \textbf{1.027 $\pm$ 0.004} & \textbf{0.493 $\pm$ 0.001} & \textbf{52.93 $\pm$ 0.14} \\
                               & cot & 0.465 $\pm$ 0.019 & 0.465 $\pm$ 0.016 & 0.320 $\pm$ 0.012 & 0.812 $\pm$ 0.017 & 1.034 $\pm$ 0.019 & \underline{0.487 $\pm$ 0.011} & \underline{52.43 $\pm$ 0.98} \\

\multirow[t]{3}{*}{Qwen2.5-72B} & zero & \underline{0.463 $\pm$ 0.003} & \underline{0.466 $\pm$ 0.006} & \underline{0.319 $\pm$ 0.005} & \underline{0.817 $\pm$ 0.009} & \underline{1.036 $\pm$ 0.003} & 0.467 $\pm$ 0.006 & 50.82 $\pm$ 0.40 \\
                               & few & 0.453 $\pm$ 0.024 & 0.459 $\pm$ 0.029 & 0.314 $\pm$ 0.021 & 0.824 $\pm$ 0.026 & 1.046 $\pm$ 0.023 & 0.445 $\pm$ 0.016 & 49.04 $\pm$ 1.30 \\
                               & cot & 0.435 $\pm$ 0.017 & 0.443 $\pm$ 0.019 & 0.302 $\pm$ 0.013 & 0.846 $\pm$ 0.022 & 1.063 $\pm$ 0.016 & 0.451 $\pm$ 0.023 & 49.55 $\pm$ 1.83 \\

\multirow[t]{3}{*}{Command-R-104B} & zero & 0.365 $\pm$ 0.041 & 0.362 $\pm$ 0.044 & 0.246 $\pm$ 0.031 & 0.876 $\pm$ 0.029 & 1.127 $\pm$ 0.037 & 0.423 $\pm$ 0.011 & 47.07 $\pm$ 0.95 \\
                               & few & 0.392 $\pm$ 0.012 & 0.381 $\pm$ 0.014 & 0.260 $\pm$ 0.012 & 0.864 $\pm$ 0.015 & 1.103 $\pm$ 0.011 & 0.406 $\pm$ 0.040 & 45.64 $\pm$ 3.21 \\
                               & cot & 0.354 $\pm$ 0.027 & 0.357 $\pm$ 0.025 & 0.242 $\pm$ 0.019 & 0.885 $\pm$ 0.020 & 1.136 $\pm$ 0.023 & 0.428 $\pm$ 0.032 & 47.42 $\pm$ 2.58 \\
\midrule
\multicolumn{2}{c}{Majority Baseline} & 0.103  $\pm$ 0.029 & 0.121  $\pm$ 0.028 & 0.081  $\pm$ 0.019 & 1.023 $\pm$ 0.010 & 1.339 $\pm$ 0.021 & 0.427 $\pm$ 0.014 & 46.87 $\pm$ 1.37 \\
\multicolumn{2}{c}{Random Baseline} & -0.021 $\pm$ 0.024 & -0.014 $\pm$ 0.029 & -0.009 $\pm$ 0.019 & 1.127 $\pm$ 0.017 & 1.429 $\pm$ 0.017 & 0.286 $\pm$ 0.005 & 35.83 $\pm$ 0.26 \\
\bottomrule
\end{tabular}}
\caption{LLM and prompting configurations scores and annotations compared against human judgment. Scores were averaged across logic, rhetoric, and dialectic dimensions, with standard deviations ($\pm$) reported across them. Pearson, Spearman, and Kendall coefficients measure correlation between LLM BT rankings and human BT rankings. MAE and RMSE measure LLM-human BT ranking differences, where lower values are better. Cohen’s $\kappa$ and percent agreement (\%) compare LLM annotations directly rather than BT scores. The best configuration per column is in \textbf{bold}; the second-best is \underline{underlined}.}
\label{tab:llm-vs-experts}
\end{table*}

\bo{We evaluate 12 open-weight models from five families: Mistral, Llama 3*, Qwen2.5, Olmo2, and Command-R, spanning parameter scales from 7B to 104B \footnote{The specific version of each LLM used in this study is provided in Appendix \ref{sec:appendix-llms-version}.}. Each family includes models of different sizes, enabling a controlled analysis of the effects of both model scale and architecture on output variability and inter-model agreement. We focus exclusively on open-weight models to ensure reproducibility and transparency.}

We evaluate whether LLMs can replicate human-derived argument quality rankings published for the Webis-ArgQuality-20 corpus. 
The approach follows the original pairwise framework 
\bo{where each comparison task 
presents the two arguments—named \emph{Argument A} and \emph{Argument B}--and instructs the LLM to decide which argument is better with respect to a single quality dimension. The response is restricted to the following: \texttt{``A''} if A is better, \texttt{``B''} if B is better, or \texttt{ ``Tie''} to indicate a tie. No additional text is allowed.}




\bo{We employed three prompting strategies: zero-shot, few-shot, and chain-of-thought. Zero-shot takes as input the dimension to judge 
and the pairwise comparison to assess. Chain-of-thought extends the zero-shot strategy by instructing the LLM to reason internally over sub-dimensions associated with each main dimension, following the framework of \citet{wachsmuth2017computational} described in Section \ref{sec:dimensions}. Logical quality is assessed through local acceptability, local relevance, and local sufficiency; rhetorical quality through credibility, emotional appeal, clarity, appropriateness, and arrangement; and dialectical quality through acceptability, relevance, and sufficiency. Each sub-dimension is reformulated as a comparative question asking which of the two arguments exhibits more of that property, accompanied by a brief definition. The model outputs only the final label for the target dimension, not intermediate sub-dimension judgments, as evaluation is conducted at the dimension level. Thus, the sub-dimensions function as structured guidance for the model's internal reasoning process rather than as prediction targets in their own right. Finally, few-shot uses the same definitions as zero-shot but dynamically adds, for the pair of arguments under evaluation, three annotated examples within their topic: one where ``\texttt{A}'' is chosen, one where ``\texttt{B}'' is chosen, and one labeled as ``\texttt{Tie}''. These examples remain the same across all comparisons within that topic.}

Specifically, we consider arguments \emph{A} and \emph{B}, and for the label \emph{A}, we select the pair with the largest score difference where the score of \emph{A} is higher than that of \emph{B}. Similarly, we choose an example for \emph{B} by selecting the pair with the largest score difference where the score of \emph{B} is higher than that of \emph{A}. For the \emph{Tie} example, we select, from the ground-truth comparisons labeled as \emph{Tie}, the pair with the smallest quality score difference for that dimension and topic. In this way, we ensure that the few-shot examples are clear-cut and avoid ambiguous examples \footnote{Each of the prompts is provided in Appendix \ref{sec:appendix-prompts}.}.


Regarding temperature, values in the 0–1 range show no statistically significant differences in performance, while values above 1 degrade it~\citep{renze-2024-effect}. We set temperature to 1 to balance stability and stochasticity for measuring intra-model reliability, with gains expected to generalize to lower settings~\citep{renze-2024-effect,LI2025242}.

Given the probabilistic nature of LLMs
, we evaluated each model across three independent runs using the same pairing strategy to measure prediction variability. The final label was determined by majority vote. The argument (A or B) with the highest frequency was selected; in cases where no single argument secured a majority (e.g., a three-way split or a tie involving an explicit 'Tie' model response), the final label was Tie. Figure~\ref{fig:pipeline} shows the overall ranking pipeline. First, the LLM evaluates each pairwise comparison three independent times, choosing between “A,” “B,” or “Tie.” Majority voting is then used to obtain a single label for each pair. Once all 13952 pairs are annotated, the resulting labels are used to fit the Bradley--Terry model and obtain rankings for the 1271 arguments in the dataset.


\begin{figure*}[t]
    \centering
    \includegraphics[width=0.87\textwidth]{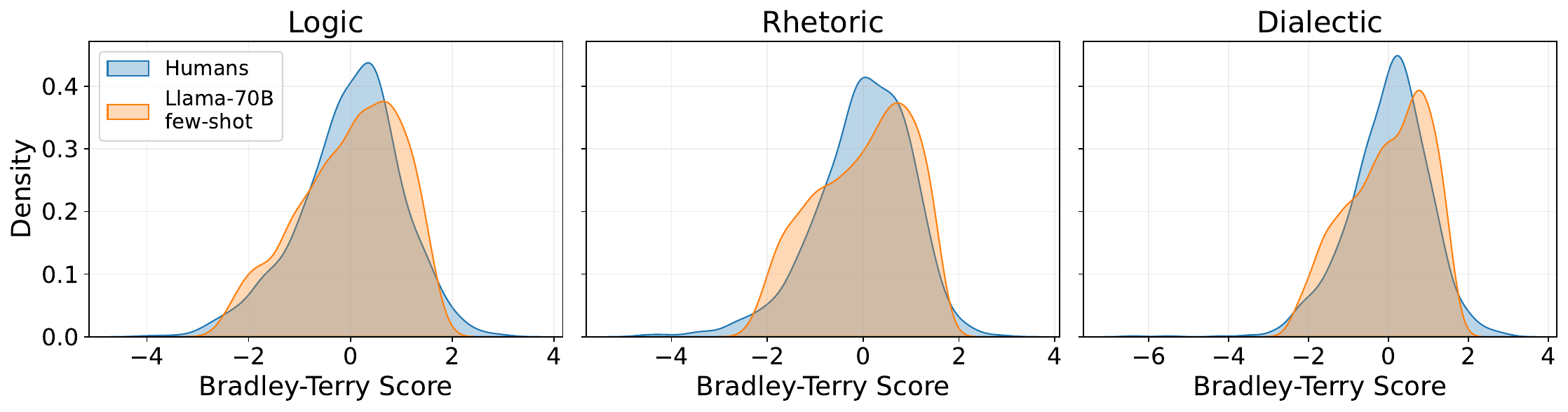}
    \caption{
    Distribution of Bradley-Terry scores assigned by humans and Llama-70B (few-shot).
    }
    \label{fig:best_llm_vs_experts_distplot}
\end{figure*}

\begin{table*}[t]
\centering
\small
\begin{tabular}{llccccccc}
\toprule
Dim & Model & Pearson ($r$) & Spearman ($\rho$) & Kendall & MAE & RMSE & $\kappa$ IAA & \% Agreement \\
\midrule
\multirow[t]{3}{*}{Logic}  & Random Baseline    & -0.041 & -0.027 & -0.018 & 1.139 & 1.443 & 0.288 & 35.64     \\
          & Majority Baseline  & 0.125  & 0.153  & 0.103  & 1.014 & 1.323 & 0.442 & 48.28     \\
          & Llama-70B Few-shot & 0.474  & 0.479  & 0.329  & 0.806 & 1.026 & 0.493 & 53.00     \\
\midrule
\multirow[t]{3}{*}{Rhetoric}   & Random Baseline    & 0.006  & 0.019  & 0.013  & 1.107 & 1.410 & 0.281 & 35.72     \\
          & Majority Baseline  & 0.071  & 0.104  & 0.069  & 1.033 & 1.363 & 0.414 & 45.55     \\
          & Llama-70B Few-shot & 0.476  & 0.488  & 0.333  & 0.808 & 1.024 & 0.493 & 52.76    \\
\midrule
\multirow[t]{3}{*}{Dialectic}  & Random Baseline    & -0.029 & -0.034 & -0.023 & 1.135 & 1.434 & 0.290 & 36.12     \\
          & Majority Baseline  & 0.113  & 0.104  & 0.070  & 1.021 & 1.332 & 0.426 & 46.78     \\
          & Llama-70B Few-shot & 0.469  & 0.464  & 0.318  & 0.818 & 1.031 & 0.494 & 53.02     \\
\bottomrule
\end{tabular}
\caption{Results per dimension of the best performing configuration LLama-70B with few-shot.}
\label{tab:best-llm-prompt}
\end{table*}

\subsection{Evaluation metrics}
To assess the reliability and consistency of LLM-based argument quality evaluation, we did three complementary analyses. 
First, in the \textbf{LLM vs. Human} analysis, we compared LLM-generated pairwise preferences and the resulting Bradley-Terry scores against those derived from the Webis-ArgQuality-20 human annotations across the logic, rhetoric, and dialectic dimensions. Second, in the \textbf{Best LLM vs. Other LLMs} analysis, we selected the highest-performing model and compared its pairwise annotations and resulting BT scores against those produced by other LLMs across the same argument pairs. Third, in the \textbf{Prediction Variability} analysis, we assessed the stability of each LLM by repeating the evaluation process across three independent runs and measuring the agreement between the resulting pairwise annotations.

\bo{We evaluated Bradley-Terry score alignment using Pearson ($r$), Spearman ($\rho$), Mean Absolute Error (MAE), and Root Mean Square Error (RMSE). Annotation agreement was measured using Cohen’s $\kappa$, Krippendorff’s $\alpha$, and percent agreement.}

\bocr{To provide reference values for the evaluation metrics, we defined two baselines: the Random Baseline, which randomly selects between "A", "B", and "Tie" for each comparison, and the Majority Baseline, which always predicts the label most frequently chosen by the human annotators, namely A for the logic, rhetoric, and dialectic dimensions (see data distribution in Appendix~\ref{sec:dist-humans-llms}). We then derived the corresponding BT scores from the comparisons produced by these baselines.}

\section{Results \& Discussion}
\label{s:execution}


\begin{table*}
\scriptsize
\centering
\resizebox{!}{5.31cm}{\
\begin{tabular}{llccccccc}
\toprule
Model & Prompt & Pearson ($r$) & Spearman ($\rho$) & Kendall & MAE & RMSE & $\kappa$ IAA & \% Agreement \\
\midrule
\multirow[t]{3}{*}{Mistral-7B} & zero & 0.521 $\pm$ 0.032 & 0.531 $\pm$ 0.027 & 0.370 $\pm$ 0.020 & 0.787 $\pm$ 0.022 & 0.978 $\pm$ 0.033 & 0.234 $\pm$ 0.089 & 26.32 $\pm$ 8.30 \\
 & few & 0.057 $\pm$ 0.242 & 0.066 $\pm$ 0.249 & 0.044 $\pm$ 0.168 & 1.085 $\pm$ 0.139 & 1.366 $\pm$ 0.181 & 0.354 $\pm$ 0.150 & 39.47 $\pm$ 15.40 \\
 & cot & 0.476 $\pm$ 0.063 & 0.492 $\pm$ 0.058 & 0.340 $\pm$ 0.042 & 0.823 $\pm$ 0.034 & 1.023 $\pm$ 0.061 & 0.134 $\pm$ 0.067 & 15.95 $\pm$ 6.04 \\
\multirow[t]{3}{*}{Llama-8B} & zero & 0.497 $\pm$ 0.014 & 0.497 $\pm$ 0.021 & 0.344 $\pm$ 0.014 & 0.797 $\pm$ 0.011 & 1.003 $\pm$ 0.014 & 0.498 $\pm$ 0.035 & 53.22 $\pm$ 3.34 \\
 & few & 0.228 $\pm$ 0.089 & 0.235 $\pm$ 0.072 & 0.158 $\pm$ 0.050 & 0.991 $\pm$ 0.058 & 1.241 $\pm$ 0.071 & 0.566 $\pm$ 0.028 & 60.09 $\pm$ 2.54 \\
 & cot & 0.194 $\pm$ 0.095 & 0.196 $\pm$ 0.094 & 0.132 $\pm$ 0.064 & 1.020 $\pm$ 0.066 & 1.268 $\pm$ 0.073 & 0.388 $\pm$ 0.014 & 42.54 $\pm$ 1.37 \\
\multirow[t]{3}{*}{Olmo2-7B} & zero & -0.033 $\pm$ 0.088 & -0.035 $\pm$ 0.079 & -0.022 $\pm$ 0.053 & 1.140 $\pm$ 0.032 & 1.436 $\pm$ 0.062 & 0.328 $\pm$ 0.078 & 37.42 $\pm$ 7.29 \\
 & few & -0.064 $\pm$ 0.116 & -0.074 $\pm$ 0.110 & -0.050 $\pm$ 0.074 & 1.162 $\pm$ 0.057 & 1.457 $\pm$ 0.080 & 0.415 $\pm$ 0.077 & 46.14 $\pm$ 7.12 \\
 & cot & 0.008 $\pm$ 0.070 & 0.005 $\pm$ 0.078 & 0.004 $\pm$ 0.052 & 1.121 $\pm$ 0.033 & 1.408 $\pm$ 0.049 & 0.371 $\pm$ 0.089 & 41.62 $\pm$ 8.51 \\
\multirow[t]{3}{*}{Qwen2.5-7B} & zero & 0.640 $\pm$ 0.051 & 0.640 $\pm$ 0.052 & 0.459 $\pm$ 0.043 & 0.658 $\pm$ 0.050 & 0.847 $\pm$ 0.059 & 0.640 $\pm$ 0.023 & 66.57 $\pm$ 2.18 \\
 & few & 0.615 $\pm$ 0.057 & 0.626 $\pm$ 0.056 & 0.450 $\pm$ 0.046 & 0.667 $\pm$ 0.055 & 0.876 $\pm$ 0.066 & 0.693 $\pm$ 0.019 & 71.64 $\pm$ 1.69 \\
 & cot & 0.496 $\pm$ 0.164 & 0.494 $\pm$ 0.176 & 0.348 $\pm$ 0.132 & 0.777 $\pm$ 0.133 & 0.995 $\pm$ 0.164 & 0.542 $\pm$ 0.103 & 57.36 $\pm$ 9.65 \\
\multirow[t]{3}{*}{Command-R-7B} & zero & 0.453 $\pm$ 0.098 & 0.451 $\pm$ 0.102 & 0.311 $\pm$ 0.076 & 0.830 $\pm$ 0.085 & 1.044 $\pm$ 0.093 & 0.505 $\pm$ 0.055 & 54.17 $\pm$ 5.10 \\
 & few & 0.382 $\pm$ 0.203 & 0.384 $\pm$ 0.203 & 0.263 $\pm$ 0.141 & 0.875 $\pm$ 0.137 & 1.102 $\pm$ 0.176 & 0.452 $\pm$ 0.075 & 49.07 $\pm$ 6.91 \\
 & cot & 0.441 $\pm$ 0.041 & 0.438 $\pm$ 0.043 & 0.299 $\pm$ 0.031 & 0.845 $\pm$ 0.028 & 1.057 $\pm$ 0.038 & 0.419 $\pm$ 0.035 & 45.64 $\pm$ 3.17 \\
\midrule
\multirow[t]{3}{*}{Mixtral-8x7B} & zero & 0.343 $\pm$ 0.067 & 0.338 $\pm$ 0.058 & 0.231 $\pm$ 0.042 & 0.904 $\pm$ 0.043 & 1.145 $\pm$ 0.059 & 0.396 $\pm$ 0.068 & 43.35 $\pm$ 6.69 \\
 & few & 0.313 $\pm$ 0.243 & 0.327 $\pm$ 0.224 & 0.224 $\pm$ 0.155 & 0.920 $\pm$ 0.164 & 1.161 $\pm$ 0.200 & 0.491 $\pm$ 0.111 & 52.61 $\pm$ 11.22 \\
 & cot & 0.181 $\pm$ 0.138 & 0.164 $\pm$ 0.128 & 0.111 $\pm$ 0.086 & 1.010 $\pm$ 0.080 & 1.277 $\pm$ 0.107 & 0.291 $\pm$ 0.055 & 32.93 $\pm$ 6.10 \\

\multirow[t]{3}{*}{Mistral-22B} & zero & -0.050 $\pm$ 0.106 & -0.053 $\pm$ 0.100 & -0.035 $\pm$ 0.067 & 1.146 $\pm$ 0.046 & 1.448 $\pm$ 0.074 & 0.385 $\pm$ 0.080 & 43.15 $\pm$ 7.41 \\
 & few & 0.062 $\pm$ 0.025 & 0.055 $\pm$ 0.021 & 0.037 $\pm$ 0.014 & 1.092 $\pm$ 0.011 & 1.370 $\pm$ 0.018 & 0.435 $\pm$ 0.044 & 47.78 $\pm$ 4.20 \\
 & cot & -0.050 $\pm$ 0.135 & -0.055 $\pm$ 0.137 & -0.037 $\pm$ 0.093 & 1.149 $\pm$ 0.071 & 1.447 $\pm$ 0.095 & 0.369 $\pm$ 0.059 & 41.53 $\pm$ 5.33 \\

\multirow[t]{3}{*}{Olmo2-32B} & zero & 0.666 $\pm$ 0.024 & 0.659 $\pm$ 0.025 & 0.472 $\pm$ 0.021 & 0.646 $\pm$ 0.027 & 0.817 $\pm$ 0.030 & 0.500 $\pm$ 0.048 & 53.17 $\pm$ 4.77 \\
 & few & 0.605 $\pm$ 0.069 & 0.594 $\pm$ 0.079 & 0.420 $\pm$ 0.062 & 0.700 $\pm$ 0.067 & 0.886 $\pm$ 0.076 & 0.529 $\pm$ 0.042 & 56.32 $\pm$ 4.03 \\
 & cot & 0.685 $\pm$ 0.019 & 0.673 $\pm$ 0.023 & 0.486 $\pm$ 0.019 & 0.623 $\pm$ 0.019 & 0.793 $\pm$ 0.024 & 0.553 $\pm$ 0.033 & 58.40 $\pm$ 3.19 \\

\multirow[t]{3}{*}{Mixtral-8x22B} & zero & 0.552 $\pm$ 0.040 & 0.545 $\pm$ 0.044 & 0.380 $\pm$ 0.033 & 0.747 $\pm$ 0.035 & 0.946 $\pm$ 0.042 & 0.429 $\pm$ 0.094 & 46.22 $\pm$ 8.98 \\
 & few & 0.509 $\pm$ 0.112 & 0.497 $\pm$ 0.109 & 0.348 $\pm$ 0.085 & 0.775 $\pm$ 0.097 & 0.986 $\pm$ 0.117 & 0.544 $\pm$ 0.055 & 57.66 $\pm$ 5.11 \\
 & cot & 0.502 $\pm$ 0.019 & 0.513 $\pm$ 0.026 & 0.356 $\pm$ 0.019 & 0.774 $\pm$ 0.018 & 0.997 $\pm$ 0.019 & 0.499 $\pm$ 0.021 & 53.20 $\pm$ 2.30 \\
\midrule
\multirow[t]{3}{*}{Llama-70B} & zero & \textbf{0.892 $\pm$ 0.003} & \textbf{0.892 $\pm$ 0.004} & \textbf{0.718 $\pm$ 0.005} & \textbf{0.341 $\pm$ 0.003} & \textbf{0.464 $\pm$ 0.007} & \textbf{0.831 $\pm$ 0.011} & \textbf{84.45 $\pm$ 1.05} \\
 & \cellcolor{gray!30} few & \cellcolor{gray!30} 1.000 $\pm$ 0.000 & \cellcolor{gray!30} 1.000 $\pm$ 0.000 & \cellcolor{gray!30} 1.000 $\pm$ 0.000 & \cellcolor{gray!30} 0.000 $\pm$ 0.000 & \cellcolor{gray!30} 0.000 $\pm$ 0.000 & \cellcolor{gray!30} 1.000 $\pm$ 0.000 & \cellcolor{gray!30} 100.00 $\pm$ 0.00 \\
 & cot & \underline{0.857 $\pm$ 0.026} & \underline{0.861 $\pm$ 0.026} & \underline{0.677 $\pm$ 0.032} & \underline{0.396 $\pm$ 0.035} & \underline{0.533 $\pm$ 0.049} & \underline{0.795 $\pm$ 0.030} & \underline{81.06 $\pm$ 2.83} \\

\multirow[t]{3}{*}{Qwen2.5-72B} & zero & 0.829 $\pm$ 0.022 & 0.824 $\pm$ 0.021 & 0.632 $\pm$ 0.023 & 0.445 $\pm$ 0.027 & 0.584 $\pm$ 0.037 & 0.741 $\pm$ 0.023 & 76.04 $\pm$ 2.17 \\
 & few & 0.819 $\pm$ 0.023 & 0.818 $\pm$ 0.021 & 0.625 $\pm$ 0.023 & 0.456 $\pm$ 0.028 & 0.600 $\pm$ 0.037 & 0.708 $\pm$ 0.030 & 72.83 $\pm$ 2.84 \\
 & cot & 0.796 $\pm$ 0.005 & 0.793 $\pm$ 0.002 & 0.600 $\pm$ 0.004 & 0.485 $\pm$ 0.010 & 0.638 $\pm$ 0.007 & 0.691 $\pm$ 0.025 & 71.29 $\pm$ 2.36 \\

\multirow[t]{3}{*}{Command-R-104B} & zero & 0.577 $\pm$ 0.077 & 0.592 $\pm$ 0.073 & 0.423 $\pm$ 0.060 & 0.705 $\pm$ 0.067 & 0.917 $\pm$ 0.086 & 0.608 $\pm$ 0.043 & 63.65 $\pm$ 4.06 \\
 & few & 0.624 $\pm$ 0.034 & 0.629 $\pm$ 0.038 & 0.452 $\pm$ 0.033 & 0.672 $\pm$ 0.036 & 0.866 $\pm$ 0.040 & 0.593 $\pm$ 0.058 & 62.22 $\pm$ 5.59 \\
 & cot & 0.563 $\pm$ 0.044 & 0.577 $\pm$ 0.046 & 0.412 $\pm$ 0.037 & 0.712 $\pm$ 0.042 & 0.934 $\pm$ 0.047 & 0.621 $\pm$ 0.057 & 64.90 $\pm$ 5.41 \\

\bottomrule
\end{tabular}}
\caption{LLM and prompt configurations scores and annotations compared against Llama-70B few-shot. The configuration most aligned with Llama-70B few-shot per column is in \textbf{bold}; the second-most aligned is \underline{underlined}.}
\label{tab:best-llm-vs-other-llms}
\end{table*}

\begin{table*}[t]
\scriptsize
\resizebox{!}{5.33cm}{\
\begin{tabular}{llccccccc}
\toprule
Model & Prompt & \multicolumn{2}{c}{All equal} & \multicolumn{2}{c}{2 equal} & \multicolumn{2}{c}{All unequal} & kripp. $\alpha$ \\
 &  & \# & \% & \# & \% & \# & \% &  \\
\midrule
\multirow[t]{3}{*}{Mistral-7B} & zero & 13952.00 $\pm$ 0.00 & 100.00 $\pm$ 0.00 & 0.00 $\pm$ 0.00 & 0.00 $\pm$ 0.00 & 0.00 $\pm$ 0.00 & 0.00 $\pm$ 0.00 & 1.00 $\pm$ 0.00 \\
 & few & 13952.00 $\pm$ 0.00 & 100.00 $\pm$ 0.00 & 0.00 $\pm$ 0.00 & 0.00 $\pm$ 0.00 & 0.00 $\pm$ 0.00 & 0.00 $\pm$ 0.00 & 1.00 $\pm$ 0.00 \\
 & cot & 13952.00 $\pm$ 0.00 & 100.00 $\pm$ 0.00 & 0.00 $\pm$ 0.00 & 0.00 $\pm$ 0.00 & 0.00 $\pm$ 0.00 & 0.00 $\pm$ 0.00 & 1.00 $\pm$ 0.00 \\

\multirow[t]{3}{*}{Llama-8B} & zero & 13952.00 $\pm$ 0.00 & 100.00 $\pm$ 0.00 & 0.00 $\pm$ 0.00 & 0.00 $\pm$ 0.00 & 0.00 $\pm$ 0.00 & 0.00 $\pm$ 0.00 & 1.00 $\pm$ 0.00 \\
 & few & 13952.00 $\pm$ 0.00 & 100.00 $\pm$ 0.00 & 0.00 $\pm$ 0.00 & 0.00 $\pm$ 0.00 & 0.00 $\pm$ 0.00 & 0.00 $\pm$ 0.00 & 1.00 $\pm$ 0.00 \\
 & cot & 13952.00 $\pm$ 0.00 & 100.00 $\pm$ 0.00 & 0.00 $\pm$ 0.00 & 0.00 $\pm$ 0.00 & 0.00 $\pm$ 0.00 & 0.00 $\pm$ 0.00 & 1.00 $\pm$ 0.00 \\

\multirow[t]{3}{*}{Olmo2-7B} & zero & 13952.00 $\pm$ 0.00 & 100.00 $\pm$ 0.00 & 0.00 $\pm$ 0.00 & 0.00 $\pm$ 0.00 & 0.00 $\pm$ 0.00 & 0.00 $\pm$ 0.00 & 1.00 $\pm$ 0.00 \\
 & few & 13952.00 $\pm$ 0.00 & 100.00 $\pm$ 0.00 & 0.00 $\pm$ 0.00 & 0.00 $\pm$ 0.00 & 0.00 $\pm$ 0.00 & 0.00 $\pm$ 0.00 & 1.00 $\pm$ 0.00 \\
 & cot & 13952.00 $\pm$ 0.00 & 100.00 $\pm$ 0.00 & 0.00 $\pm$ 0.00 & 0.00 $\pm$ 0.00 & 0.00 $\pm$ 0.00 & 0.00 $\pm$ 0.00 & 1.00 $\pm$ 0.00 \\

\multirow[t]{3}{*}{Qwen2.5-7B} & zero & 13565.67 $\pm$ 669.15 & 97.23 $\pm$ 4.80 & 386.33 $\pm$ 669.15 & 2.77 $\pm$ 4.80 & 0.00 $\pm$ 0.00 & 0.00 $\pm$ 0.00 & 0.97 $\pm$ 0.05 \\
 & few & 13952.00 $\pm$ 0.00 & 100.00 $\pm$ 0.00 & 0.00 $\pm$ 0.00 & 0.00 $\pm$ 0.00 & 0.00 $\pm$ 0.00 & 0.00 $\pm$ 0.00 & 1.00 $\pm$ 0.00 \\
 & cot & 13952.00 $\pm$ 0.00 & 100.00 $\pm$ 0.00 & 0.00 $\pm$ 0.00 & 0.00 $\pm$ 0.00 & 0.00 $\pm$ 0.00 & 0.00 $\pm$ 0.00 & 1.00 $\pm$ 0.00 \\

\multirow[t]{3}{*}{Command-R-7B} & zero & 13952.00 $\pm$ 0.00 & 100.00 $\pm$ 0.00 & 0.00 $\pm$ 0.00 & 0.00 $\pm$ 0.00 & 0.00 $\pm$ 0.00 & 0.00 $\pm$ 0.00 & 1.00 $\pm$ 0.00 \\
 & few & 13952.00 $\pm$ 0.00 & 100.00 $\pm$ 0.00 & 0.00 $\pm$ 0.00 & 0.00 $\pm$ 0.00 & 0.00 $\pm$ 0.00 & 0.00 $\pm$ 0.00 & 1.00 $\pm$ 0.00 \\
 & cot & 13952.00 $\pm$ 0.00 & 100.00 $\pm$ 0.00 & 0.00 $\pm$ 0.00 & 0.00 $\pm$ 0.00 & 0.00 $\pm$ 0.00 & 0.00 $\pm$ 0.00 & 1.00 $\pm$ 0.00 \\
\midrule
\multirow[t]{3}{*}{Mixtral-8x7B} & zero & 13105.00 $\pm$ 188.65 & 93.93 $\pm$ 1.35 & 847.00 $\pm$ 188.65 & 6.07 $\pm$ 1.35 & 0.00 $\pm$ 0.00 & 0.00 $\pm$ 0.00 & 0.92 $\pm$ 0.01 \\
 & few & 12025.33 $\pm$ 1331.71 & 86.19 $\pm$ 9.54 & 1898.00 $\pm$ 1298.63 & 13.60 $\pm$ 9.31 & 28.67 $\pm$ 48.79 & 0.21 $\pm$ 0.35 & 0.78 $\pm$ 0.02 \\
 & cot & 12567.00 $\pm$ 543.52 & 90.07 $\pm$ 3.90 & 1379.33 $\pm$ 533.88 & 9.89 $\pm$ 3.83 & 5.67 $\pm$ 9.81 & 0.04 $\pm$ 0.07 & 0.87 $\pm$ 0.06 \\

\multirow[t]{3}{*}{Mistral-22B} & zero & 6787.00 $\pm$ 1466.48 & 48.65 $\pm$ 10.51 & 6739.67 $\pm$ 1419.55 & 48.31 $\pm$ 10.17 & 425.33 $\pm$ 61.09 & 3.05 $\pm$ 0.44 & 0.16 $\pm$ 0.04 \\
 & few & 4899.67 $\pm$ 1359.89 & 35.12 $\pm$ 9.75 & 7971.67 $\pm$ 778.22 & 57.14 $\pm$ 5.58 & 1080.67 $\pm$ 584.11 & 7.75 $\pm$ 4.19 & 0.16 $\pm$ 0.07 \\
 & cot & 5320.00 $\pm$ 999.99 & 38.13 $\pm$ 7.17 & 7830.00 $\pm$ 753.81 & 56.12 $\pm$ 5.40 & 802.00 $\pm$ 374.15 & 5.75 $\pm$ 2.68 & 0.11 $\pm$ 0.00 \\

\multirow[t]{3}{*}{Olmo2-32B} & zero & 13952.00 $\pm$ 0.00 & 100.00 $\pm$ 0.00 & 0.00 $\pm$ 0.00 & 0.00 $\pm$ 0.00 & 0.00 $\pm$ 0.00 & 0.00 $\pm$ 0.00 & 1.00 $\pm$ 0.00 \\
 & few & 11310.00 $\pm$ 4576.08 & 81.06 $\pm$ 32.80 & 2642.00 $\pm$ 4576.08 & 18.94 $\pm$ 32.80 & 0.00 $\pm$ 0.00 & 0.00 $\pm$ 0.00 & 0.81 $\pm$ 0.33 \\
 & cot & 13952.00 $\pm$ 0.00 & 100.00 $\pm$ 0.00 & 0.00 $\pm$ 0.00 & 0.00 $\pm$ 0.00 & 0.00 $\pm$ 0.00 & 0.00 $\pm$ 0.00 & 1.00 $\pm$ 0.00 \\

\multirow[t]{3}{*}{Mixtral-8x22B} & zero & 9584.00 $\pm$ 585.15 & 68.69 $\pm$ 4.19 & 4229.00 $\pm$ 527.10 & 30.31 $\pm$ 3.78 & 139.00 $\pm$ 59.77 & 1.00 $\pm$ 0.43 & 0.63 $\pm$ 0.01 \\
 & few & 6598.67 $\pm$ 847.45 & 47.30 $\pm$ 6.07 & 6379.33 $\pm$ 667.24 & 45.72 $\pm$ 4.78 & 974.00 $\pm$ 194.13 & 6.98 $\pm$ 1.39 & 0.39 $\pm$ 0.09 \\
 & cot & 7869.00 $\pm$ 645.82 & 56.40 $\pm$ 4.63 & 5667.67 $\pm$ 632.36 & 40.62 $\pm$ 4.53 & 415.33 $\pm$ 80.08 & 2.98 $\pm$ 0.57 & 0.50 $\pm$ 0.06 \\
\midrule
\multirow[t]{3}{*}{Llama-70B} & zero & 13268.00 $\pm$ 64.63 & 95.10 $\pm$ 0.46 & 684.00 $\pm$ 64.63 & 4.90 $\pm$ 0.46 & 0.00 $\pm$ 0.00 & 0.00 $\pm$ 0.00 & 0.93 $\pm$ 0.01 \\
 & few & 13952.00 $\pm$ 0.00 & 100.00 $\pm$ 0.00 & 0.00 $\pm$ 0.00 & 0.00 $\pm$ 0.00 & 0.00 $\pm$ 0.00 & 0.00 $\pm$ 0.00 & 1.00 $\pm$ 0.00 \\
 & cot & 13210.33 $\pm$ 645.63 & 94.68 $\pm$ 4.63 & 740.00 $\pm$ 643.95 & 5.30 $\pm$ 4.62 & 1.67 $\pm$ 2.89 & 0.01 $\pm$ 0.02 & 0.93 $\pm$ 0.06 \\

\multirow[t]{3}{*}{Qwen2.5-72B} & zero & 12743.33 $\pm$ 1053.37 & 91.34 $\pm$ 7.55 & 1208.67 $\pm$ 1053.37 & 8.66 $\pm$ 7.55 & 0.00 $\pm$ 0.00 & 0.00 $\pm$ 0.00 & 0.90 $\pm$ 0.09 \\
 & few & 13347.67 $\pm$ 1046.74 & 95.67 $\pm$ 7.50 & 582.00 $\pm$ 1008.05 & 4.17 $\pm$ 7.23 & 22.33 $\pm$ 38.68 & 0.16 $\pm$ 0.28 & 0.95 $\pm$ 0.09 \\
 & cot & 13433.00 $\pm$ 898.93 & 96.28 $\pm$ 6.44 & 519.00 $\pm$ 898.93 & 3.72 $\pm$ 6.44 & 0.00 $\pm$ 0.00 & 0.00 $\pm$ 0.00 & 0.96 $\pm$ 0.07 \\

\multirow[t]{3}{*}{Commandr-104B} & zero & 10567.67 $\pm$ 3030.76 & 75.74 $\pm$ 21.72 & 3256.33 $\pm$ 2878.99 & 23.34 $\pm$ 20.63 & 128.00 $\pm$ 221.70 & 0.92 $\pm$ 1.59 & 0.67 $\pm$ 0.30 \\
 & few & 12596.67 $\pm$ 2347.51 & 90.29 $\pm$ 16.83 & 1355.33 $\pm$ 2347.51 & 9.71 $\pm$ 16.83 & 0.00 $\pm$ 0.00 & 0.00 $\pm$ 0.00 & 0.86 $\pm$ 0.24 \\
 & cot & 12541.00 $\pm$ 2443.92 & 89.89 $\pm$ 17.52 & 1315.33 $\pm$ 2278.22 & 9.43 $\pm$ 16.33 & 95.67 $\pm$ 165.70 & 0.69 $\pm$ 1.19 & 0.82 $\pm$ 0.31 \\

\bottomrule
\end{tabular}}
\caption{Variability across three runs per LLM and prompting configuration in the pairwise comparison annotation task. Reported are the proportions of instances with identical labels across all three runs (all equal), agreement between two of three runs (2 equal), and different labels across all runs (unequal), along with Krippendorff’s $\alpha$. Results are averaged across dimensions ($\pm$ standard deviation).}
\label{tab:prediction-variability}
\end{table*}

\bo{To address \textbf{SRQ1}, we begin by identifying the best-performing model and prompt configuration for evaluating argument quality. Table \ref{tab:llm-vs-experts} compares rankings and annotations produced by humans with those generated by LLMs under zero-shot (zero), few-shot (few), and chain-of-thought (cot) prompting strategies, across 36 configurations comprising 12 models and 3 prompting strategies. Llama-70B with few-shot prompting achieves the highest average correlations across the Pearson, Spearman, and Kendall metrics over the logic, rhetoric, and dialect dimensions, with averaged scores of 0.473, 0.477, and 0.327, respectively. It is followed by Qwen2.5-72B under zero-shot prompting, which attains corresponding averaged scores of 0.463, 0.466, and 0.319. Both models also exhibit low variability across the three dimensions \bocr{and obtain scores well superior to random and majority baselines}. Both configurations achieve the lowest error metrics, with MAE and RMSE of 0.811 and 1.027 for LLaMA-70B with few-shot, and 0.817 and 1.036 for Qwen2.5-72B with zero-shot, respectively. The highest Cohen's $\kappa$ is obtained also by LLama-70B with few-shot with 0.493, while the second best is the same model under both zero-shot and chain-of-thought. Similarly, these configurations achieve a percent agreement of 52.93\%, 52.43\%, and 52.33\% for few-shot, chain-of-thought, and zero-shot, respectively.} 
\bocr{To test for statistical significance, we run a one-tailed bootstrap test (N=10000) across all dimensions and metrics. At the 95\% confidence level, Llama-70B few-shot is significantly better than all other configurations, except Llama-70B with chain-of-thought and Qwen-2.5-72B with zero and few-shot.}

\bocr{Note also that, Olmo2-7B and Mistral-22B, obtain negative correlation values, although these values remain close to zero. Nevertheless, both models correctly achieve a moderate IAA with respect to the ground truth. Olmo2-7B achieves an average Cohen's Kappa ranging from 0.331 to 0.401 across the quality dimensions, while Mistral-22B achieves an average Cohen's Kappa ranging from 0.366 to 0.381. Looking at the comparison distributions (Table~\ref{tab:human-llm-comp-dist} in Appendix~\ref{sec:dist-humans-llms}), both Olmo2-7B and Mistral-22B predict argument B more often across pairwise comparisons, regardless of the prompting configuration or quality dimension. Since B is the preferred label in approximately 40\% of the ground-truth comparisons, this tendency can result in moderate inter-annotator agreement. However, these predictions are not sufficiently informative for BT because they provide little discrimination between arguments. As a result, the models produce BT rankings that correlate poorly with the ground-truth rankings, leading to negative values.}

\noindent \textbf{Performance by model size}. Small and medium size models do not exhibit higher agreement with humans than large size ones. Across prompts, chain-of-thought may lead to lower correlation scores and agreement with human rankings and annotations compared to zero-shot. Overall, zero-shot, few-shot, and chain-of-thought exhibit minimal variability across the evaluated dimensions.

\noindent \textbf{Performance by model family}. 
Command-R-7B and Command-R-104B achieve comparable performance, with Command-R-104B obtaining slightly higher correlation and agreement scores. In most other model families, smaller variants perform worse than their larger counterparts. The main exception is the Mistral family, where Mistral-7B achieves higher correlation with human rankings across all prompting configurations than Mistral-22B, while also obtaining lower MAE and RMSE values. 

\noindent \textbf{Performance by prompt strategy}. There is no systematic preference for a single prompting strategy across all models. Among small models, except for Qwen2.5-7B, few-shot prompting decreases correlation and agreement with humans relative to zero-shot prompting. A similar pattern is observed in medium-sized models, where few-shot prompting is generally detrimental, except for Mixtral-8x7B and Mixtral-8x22B. In contrast, among large models, few-shot prompting benefits Llama-70B and Command-R-104B.

\noindent \textbf{Performance by quality dimension}. Overall, configurations showed low variability across the logical, quality, and rhetorical dimensions, indicating that no specific dimension was harder to evaluate. Only Mixtral-8x7B and Command-R-7B under the few-shot setting presented higher variability compared to other configurations across Pearson, Spearman, Kendall, MAE, and RMSE. However, this variability was not reflected in Cohen's $\kappa$ or percent agreement. Table \ref{tab:best-llm-prompt} shows the highest performance achieved by the best configuration, Llama-70B with few-shot, highlighting that the three dimensions had similar difficulty levels \footnote{The full table of results for each model, prompt, and dimension can be found in the repository of the paper.}. Moreover, Figure \ref{fig:best_llm_vs_experts_distplot} shows that, despite its performance, Llama-70B with few-shot produces a BT score distribution that differs from that of humans, consistent with the moderate correlation observed. In particular, the LLM's distribution appears slightly more right-skewed.

\subsection{Cross-LLM Comparison}

\bo{In order to address \textbf{SRQ2}, Table \ref{tab:best-llm-vs-other-llms} compares the rankings and annotations produced by Llama-70B with few-shot against those generated by the remaining models. While configurations other than Llama-70B with few-shot show lower results with respect to the ground truth, we also evaluate how closely other LLMs agree with this best configuration.}

\bo{Among all configurations, Llama-70B consistently shows the highest correlations and agreement scores across zero-shot and chain-of-thought. For zero-shot it achieves 0.892, 0.892, and 0.718 for Pearson, Spearman, and Kendall, respectively, while 0.831 and 84.45\% for Cohen's $\kappa$ and percent agreement, respectively. Qwen2.5-72B follows closely under all prompting settings. Although Command-R-104B still shows substantial agreement, it consistently performs below the other large models, Llama-70B and Qwen2.5-72B.}

\bo{Within the medium-sized models, Olmo2-32B is the closest to the best configuration. It surpasses Command-R-104B under zero-shot and chain-of-thought, while falling slightly behind under few-shot. The remaining medium-sized models show only moderate agreement. An exception is Mistral-22B, which obtains correlation scores close to 0 together with higher MAE and RMSE values.}

\bo{The small models generally show moderate agreement and correlation scores. Qwen2.5-7B is the closest among the small models, reflecting the strong results of its larger 72B counterpart across all prompting settings. In contrast, Llama-8B achieves only moderate IAA and moderate-to-low correlation scores, together with higher MAE and RMSE values compared to the models with the strongest agreement. Similarly, Olmo2-7B does not align as well with the best configuration as its 32B counterpart, producing correlations close to 0. Command-R-7B, however, remains partially aligned as its larger version. The low variability across dimensions indicates that these findings are consistent for each of them
\footnote{The full results table across models, prompts, and dimensions can be found in the repository of the paper.}.}

These results, together with those in Table~\ref{tab:llm-vs-experts}, show that although Llama-70B is the best configuration and reaches moderate agreement, some groups of LLMs closely align with it and \textit{share its vision}, while others are less aligned with it despite still agreeing with humans
, reflecting differing but potentially accurate interpretations of the dimensions.


\subsection{Prediction Variability}

\bo{In order to address \textbf{SRQ3}, Table \ref{tab:prediction-variability} shows how LLMs exhibit low prediction variability in the pairwise comparison annotation task. The worst-case scenario is Mistral-22B with few-shot prompting, where 7.75\% of cases resulted in all three possible labels (``\texttt{A}'', ``\texttt{B}'', and ``\texttt{Tie}'') across runs. Mistral-22B also achieved the lowest Krippendorff's $\alpha$ scores overall (0.16, 0.16, and 0.11 for zero-shot, few-shot, and chain-of-thought, respectively). This is followed by Mixtral-8x22B, which obtained Krippendorff's $\alpha$ scores of 0.63, 0.39, and 0.50 under zero-shot, few-shot, and chain-of-thought, respectively. Among the less stable models, Command-R-104B achieved $\alpha$ values of 0.67 and 0.82 with zero-shot and chain-of-thought, respectively, which improved to 0.86 with few-shot. All remaining models and prompting strategies exhibited strong agreement, achieving Krippendorff's $\alpha$ values above 0.78.}

\bo{Although small models do not achieve the highest performance relative to humans, they are generally the most stable in the pairwise comparison annotation task. They consistently attain perfect or almost perfect, Krippendorff's $\alpha$ scores, while producing few or no "2 equal" and "All unequal" cases. Large models show slightly lower stability. Still, benefiting from few-shot prompting, resulting in fewer "All unequal", "2 equal" cases, or better inter-annotator agreement. In contrast, few-shot prompting appears detrimental for medium-sized models, which consistently achieve lower Krippendorff's $\alpha$ scores than their zero-shot and chain-of-thought counterparts.}

\section{Conclusion \& Future Work}\label{s:conclusion}


\bo{This paper investigated the potential of 12 open-weight LLMs to evaluate argument quality across the logical, rhetorical, and dialectical dimensions using zero-shot, few-shot, and chain-of-thought. Through pairwise comparisons and Bradley–Terry modeling, we analyzed both the alignment of LLM-generated rankings with human judgments and their consistency across runs.}

\bo{We show that, although LLMs do not yet reach human-level performance, they can approximate BT rankings derived from human pairwise comparisons to a meaningful degree. Among all evaluated models, Llama-70B with few-shot prompting achieved the strongest results, obtaining the highest (but moderate) agreement with human annotations, with Pearson, Spearman, and Kendall correlations of 0.473, 0.477, and 0.327, respectively, as well as a Cohen’s $\kappa$ of 0.493 and 52.93 percent agreement, across the three quality dimensions. 
Some models align more closely with Llama-70B, such as Qwen-2.5-72B, or smaller models like Qwen-2.5-7B and Olmo2-32B, while others are less consistent with it but still agree with humans. This suggests that different models may capture complementary perspectives, and that combining diverse LLMs may outperform relying on Llama-70B alone.}

\bo{Finally, we show that, across only three runs per model, prediction variability remains low. Disagreements ranged from cases where two out of three labels matched to instances where all labels differed (up to 7.75\% of the total comparisons for the worst configuration Mistral-22B). While still low, this variability motivates the use of aggregation strategies such as majority voting. Few-shot prompting was also found to enhance stability, although its benefits were primarily observed in larger models.}

\bo{Future work will explore fine-tuning~\cite{stahl-etal-2025-arginstruct} and ensembling LLMs for argument quality assessment, including dimension-specific models. Domain-specific training and ensembling may improve their ability to evaluate reasoning validity and detect fallacies.} 
\bo{Second, we will broaden the scope of argument quality assessment by incorporating additional dimensions and working toward a comprehensive overall quality measure.} 
\section{Limitations}\label{s:threats}

In this section, the limitations of the study are discussed as well as the potential threats to the validity of the findings. The discussion is structured according to the common validity categories inspired by~\citet{wohlin2012} guidelines, namely internal, external, construct and conclusion validity. 

\subsection{Internal Validity}

One limitation of this study is the inherent variability of LLM-based evaluations, as different runs may produce different pairwise judgments. While the results remained stable across runs and majority voting was used to improve reliability, stability itself remains an important factor to assess. In particular, we did not evaluate the models at lower temperature settings. Based on the hypothesis proposed by~\citet{renze-2024-effect}, which identifies a plateau beyond which model outputs may become unreliable, stability at a temperature of 1 may indicate comparable or greater stability at lower temperatures. Nevertheless, an ablation study across different temperature settings would be necessary to verify this assumption and confirm that all models produce comparable results for our task in line with the findings of~\citet{renze-2024-effect}.

Another limitation concerns prompt formulation. Although the same prompt templates were used for all models, differences in how individual LLMs interpret the instructions may have influenced the resulting judgments.


\subsection{External Validity}
First, the evaluation was based on a specific dataset consisting of arguments drawn from an online debate portal on a set of different topics. These arguments are mostly a few sentences. Therefore, the results achieved may not generalize to other more complex forms, such as essays. There is a possibility that it could be (more or less) challenging for the LLM. 

Second, the scores used as ground truth came from a specific annotation process and reflect the judgments of the annotators shaped by their own interpretation of the dimensions. In another context, the different dimensions could be understood differently, meaning the LLM's alignment with one set of humans might not hold if used in another set. 

Third, zero-shot, \bo{few-shot, and chain-of-thought} approaches were used where the model is given instructions to compare the arguments without fine-tuning or additional training. While these approaches are practical, it does not take into account the performance gains from fine-tuning the LLM. 
The results do not explain how well the models would do in that particular setting. 

\bocr{Fourth, while the Webis-ArgQuality-20 corpus has been used to evaluate LLMs in recent studies of computational argumentation, it is difficult to reliably assess the extent of potential data contamination as the training data of most modern LLMs is not publicly disclosed, being a current issue in LLM-based studies. Among the 12 models evaluated, only the OLMo family provides full transparency regarding its training data. The remaining open-weight models (Llama, Qwen, Command-R, and Mistral) do not disclose their training datasets or sources (except Llama that reports using public online data pre-December-2023).}

Finally, LLMs are evolving rapidly. The LLM versions used in this study are not the latest and may have already been surpassed by newer releases. At the same time, to ensure a fair comparison, we accounted for model size (small, medium, and large) and family, while avoiding substantial version differences (e.g., comparing a version 3 model with a version 1 model from the same family). Although comparative and qualitative insights might remain relevant, metric values may shift even further with advances in LLM.
\subsection{Construct Validity}
Argument quality was split into three dimensions (logical, rhetorical, dialectical) according to the dataset's scheme for annotation. There is a possibility that the LLMs comparisons do not align with the intended construct. For example the LLM might rate the argument's logical quality on confident phasing even if the actual argument itself is logically flawed. 
Comparing them to human judgment without 100\% certainty that they have the same understanding of the 'quality' dimensions is very challenging. 
Furthermore, while this approach provided insights that were valuable, it did not encompass every aspect related to argument quality such as precision and completeness. 
The dataset used for the gold standard comes from different human crowdsourcers who made pairwise comparisons. Any noise and/or bias could affect the results. The Webis-ArgQuality dataset is assumed as a valid measure for argument quality, but it is possible that there could have been inconsistencies in the annotation process or maybe the annotators had their own biases. Another consideration is the separation of quality into three dimensions. Each dimension is treated independently for evaluation purposes, but there was no explicit reminder in the prompt to ignore the other dimensions. 
Therefore, there is a possibility that the LLM might try to bring in other dimensions despite the intention to isolate. If the LLM for example took rhetorical into account for logical dimension analysis, then results could be blurry.  
\subsection{Conclusion Validity}
One limitation related to conclusion validity is seen through the number of pairwise comparisons generated for this study. Although 13952 
comparisons across 1271 arguments provided a substantial dataset, prior work by~\citet{gienapp-etal-2020-efficient} demonstrated the benefits of using far larger comparisons sets (over 41000). Larger datasets may increase the statistical robustness of the Bradley-Terry model, and improve the stability of resulting rankings. \bo{In this study, the comparison set was chosen to maintain consistency with the comparison corpus released as part of the Webis-ArgQuality-20, which relies on a cyclic group sampling strategy. Future research could explore the impact of scaling up the number of comparisons.} 

\section{Ethical Considerations}
The assessment of argument quality aims, in general, at the societal benefit since better, stronger, higher-quality arguments do not contain fallacies and spurious reasoning. Therefore, this work intends to promote the development of high-quality, grounded discussions.
At the same time, two ethical risks are inherent to this work. The first is that, by strengthening the quality of arguments supporting them, unethical or harmful claims are spread. For instance, hateful messages could be supported by rhetorically high-quality arguments. One possible solution to this issue is to accompany argument quality assessment with hate speech detection or similar analyses involving other information quality aspects.
The second risk is that, by demanding quality assessment to LLMs, intentional or unintentional distortions are introduced in the analysis and moderation of public discourse. For example, models could show biases that favor some types of arguments over others. For this reason, as indicated also above, it is important to include human oversight in these moderation systems and employ automated models in LLM-in-the-loop systems.

\section*{Acknowledgments}

This work used the Dutch national e-infrastructure with the support of the SURF Coop-erative using grant no. EINF-15500. The authors would also like to thank Greta Damo for her helpful comments during the revision of this paper.
\bibliography{custom}

\vfill \break 
\clearpage
\newpage
\appendix

\setcounter{table}{0}
\renewcommand{\tablename}{Table} 
\renewcommand{\thetable}{\Alph{table}}


\section{Specific Version of LLMs Evaluated}
\label{sec:appendix-llms-version}
We use Mistral~\cite{jiang2023mistral7b}, Llama 3*~\cite{dubey2024llama}, Qwen2.5~\cite{qwen2.5}, Olmo2~\cite{olmo20252olmo2furious}, and Command-R~\cite{cohere2025commandaenterprisereadylarge} open-access Instruct-based models from Hugging Face \footnote{\url{https://huggingface.co/models}}. All models are publicly available and traceable through their repository identifiers:

\noindent \textbf{Small Size Models}:
\begin{itemize}
    \item Mistral-7B: \texttt{mistralai/} \\ \texttt{Mistral-7B-Instruct-v0.3}
    \item Llama-8B: \texttt{meta-llama/} \\ \texttt{Llama-3.1-8B-Instruct}
    \item Olmo2-7B: \texttt{allenai/} \\ \texttt{OLMo-2-1124-7B-Instruct}
    \item Qwen2.5-7B: \texttt{Qwen/Qwen2.5-7B-Instruct}
    \item Command-r-7B: \texttt{CohereLabs/}\texttt{c4ai-command-r7b-12-2024}
\end{itemize}

\noindent \textbf{Medium Size Models}:
\begin{itemize}
    \item Mixtral-8x7B: \texttt{mistralai/} \\ \texttt{Mixtral-8x7B-Instruct-v0.1}
    \item Mistral-22B: \texttt{mistralai/} \\ \texttt{Mistral-Small-Instruct-2409}
    \item Olmo2-32B: \texttt{allenai/} \\ \texttt{OLMo-2-0325-32B-Instruct}
    \item Mixtral-8x22B: \texttt{mistralai/} \\ \texttt{Mixtral-8x22B-Instruct-v0.1}
\end{itemize}

\noindent \textbf{Large Size Models}:
\begin{itemize}
    \item Llama-70B: \texttt{meta-llama/} \\ \texttt{Llama-3.3-70B-Instruct}
    \item Qwen2.5-72B: \texttt{Qwen/} \\ \texttt{Qwen2.5-72B-Instruct}
    \item Command-r-104B: \texttt{CohereLabs/} \\ \texttt{c4ai-command-r-plus-08-2024}
\end{itemize}

Small models were run on a single H100 GPU, medium-sized models on up to two H100 GPUs, and large models on up to four H100 GPUs.

\section{Prompts} 
\label{sec:appendix-prompts}

We tested three prompting strategies, zero-shot, few-shot, and chain-of-thought using a system-user prompt setup with structured outputs through the vLLM Python library \footnote{\url{https://docs.vllm.ai/en/latest/}}.

\subsection{Zero-shot Prompt}

\noindent \textbf{System Prompt}: \texttt{You are given two arguments: Argument A and Argument B. Decide which one is \{dim\_adverb\} stronger: Reply with only one of the following options: A, B, or Tie.}

\noindent \textbf{User Prompt}: \\ \texttt{Argument A: \{a\} \\ Argument B: \{b\}}

\noindent \textbf{Output Labels}: \texttt{["A", "B", "Tie"]}

Where \texttt{dim\_adverb} are replaced with the strings \texttt{logically}, \texttt{dialectically}, \texttt{rhetorically}, for the logic, dialectic, and rhetoric dimension, respectively. \texttt{a}, and \texttt{b} are replaced with the arguments A, and B, respectively, to be judged by the LLM.

\subsection{Few-shot Prompt}

\noindent \textbf{System Prompt}: \texttt{You are given two arguments: Argument A and Argument B. Decide which one is \{dim\_adverb\} stronger: Reply with only one of the following options: A, B, or Tie. \\ \\
Examples: \\
Argument A: \{ex\_A\_argA\} \\
Argument B: \{ex\_A\_argB\} \\
A \\ \\
Argument A: \{ex\_B\_argA\} \\
Argument B: \{ex\_B\_argB\} \\
B \\ \\
Argument A: \{ex\_tie\_argA\} \\
Argument B: \{ex\_tie\_argB\} \\
tie}

\noindent \textbf{User Prompt}: \\ \texttt{Argument A: \{a\} \\ Argument B: \{b\}}

\noindent \textbf{Output Labels}: \texttt{["A", "B", "Tie"]}

Where \texttt{dim\_adverb}, \texttt{a}, and \texttt{b} are instantiated as in the zero-shot prompt. \texttt{ex\_A\_argA} and \texttt{ex\_A\_argB} are few-shot examples where \texttt{A} is preferred on the evaluated dimension; \texttt{ex\_B\_argA} and \texttt{ex\_B\_argB} are examples where \texttt{B} is preferred; and \texttt{ex\_tie\_argA} and \texttt{ex\_tie\_argB} are examples where the preferred label is \texttt{Tie}.

 We used 3 shots for each of the 20 topics in the Webis-ArgQuality-20 corpus, leading to 60 examples per dimension \footnote{The full list of selected examples is available at \url{https://github.com/benjaminocampo/arg_quality_with_llms}.}.

\subsection{Chain-of-thought}

\noindent \textbf{System Prompt}: \texttt{You are given two arguments: Argument A and Argument B. \\
Decide which one is \{dim\_adverb\} stronger considering these criteria: \\
\{dim\_criterias\} \\
Let's think step by step and reply with only one of the following options: A, B, or Tie.}

\noindent \textbf{User Prompt}: \\ \texttt{Argument A: \{a\} \\ Argument B: \{b\}}

\noindent \textbf{Output Labels}: \texttt{["A", "B", "Tie"]}

Where \texttt{dim\_adverb}, \texttt{a}, and \texttt{b} are instantiated as in the zero-shot prompt. \texttt{dim\_criterias} is replaced with the chain-of-thought questions for each quality dimension based on the theoretical work of~\citet{wachsmuth2017computational}:

\noindent \textbf{Criteria for logic}: \\
\texttt{- which is more acceptable? A premise of an argument is acceptable if it is rationally worthy of being believed to be true. \\
- which is more relevant to a conclusion? A premise of an argument is relevant if it contributes to the acceptance or rejection of the argument's conclusion. \\
- which is more sufficient to justify a conclusion? An argument's premises are sufficient if, together, they give enough support to make it rational to draw its conclusion.}

\noindent \textbf{Criteria for dialectic}: \\
\texttt{- which would be acceptable to the audience? Argumentation is acceptable if the target audience accepts both the consideration of the stated arguments for the issue and the way they are stated. \\
- which contributes more to constructive dialogue? Argumentation is relevant if it contributes to the issue's resolution, i.e., if it states arguments or other information that help to arrive at an ultimate conclusion. \\
- which better anticipates or refutes counterarguments? Argumentation is sufficient if it adequately rebuts those counterarguments to it that can be anticipated.}

\noindent \textbf{Criteria for rhetoric}: \\
\texttt{- which appears more authorative/trust worthy? Argumentation creates credibility if it conveys arguments and similar in a way that makes the author worthy of credence. \\
- which makes a stronger emotional appeal? Argumentation makes a successful emotional appeal if it creates emotions in a way that makes the target audience more open to the author's arguments. \\
- which has a clearer style? Argumentation has a clear style if it uses correct and widely unambiguous language as well as if it avoids unnecessary complexity and deviation from the issue. \\
- which has a more appropiate style? Argumentation has an appropriate style if the used language supports the creation of credibility and emotions as well as if it is proportional to the issue. \\
- Which is better arranged? Argumentation is arranged properly if it presents the issue, the arguments, and its conclusion in the right order.}

\section{Licenses and Intended Use}

\begin{table*}[t]
\small
\centering
\resizebox{!}{4.7cm}{\
\begin{tabular}{llrrrrrr|rrrrrr|rrrrrr}
\toprule
 &  & \multicolumn{6}{c|}{Logic} & \multicolumn{6}{c|}{Rhetoric} & \multicolumn{6}{c}{Dialectic} \\
 &  & \multicolumn{2}{c}{Zero} & \multicolumn{2}{c}{Few} & \multicolumn{2}{c|}{CoT} & \multicolumn{2}{c}{Zero} & \multicolumn{2}{c}{Few} & \multicolumn{2}{c|}{CoT} & \multicolumn{2}{c}{Zero} & \multicolumn{2}{c}{Few} & \multicolumn{2}{c}{CoT} \\
 &  & \# & \% & \# & \% & \# & \% & \# & \% & \# & \% & \# & \% & \# & \% & \# & \% & \# & \% \\
\midrule
\multirow[t]{3}{*}{Mistral-7B} & A & 1265 & 9.07 & 605 & 4.34 & 458 & 3.28 & 4163 & 29.84 & 11901 & 85.30 & 1274 & 9.13 & 2776 & 19.90 & 3432 & 24.60 & 2075 & 14.87 \\
                                     & B & 1443 & 10.34 & 13204 & 94.64 & 650 & 4.66 & 2057 & 14.74 & 1860 & 13.33 & 1144 & 8.20 & 1445 & 10.36 & 1734 & 12.43 & 1553 & 11.13 \\
                                     & Tie & 11244 & 80.59 & 143 & 1.02 & 12844 & 92.06 & 7732 & 55.42 & 191 & 1.37 & 11534 & 82.67 & 9731 & 69.75 & 8786 & 62.97 & 10324 & 74.00 \\
\cline{2-20}
\multirow[t]{3}{*}{Llama-8B} & A & 6246 & 44.77 & 9038 & 64.78 & 6407 & 45.92 & 7401 & 53.05 & 10510 & 75.33 & 7601 & 54.48 & 5891 & 42.22 & 11318 & 81.12 & 5645 & 40.46 \\
                                     & B & 4663 & 33.42 & 4617 & 33.09 & 2687 & 19.26 & 4868 & 34.89 & 3372 & 24.17 & 3027 & 21.70 & 5470 & 39.21 & 2589 & 18.56 & 3768 & 27.01 \\
                                     & Tie & 3043 & 21.81 & 297 & 2.13 & 4858 & 34.82 & 1683 & 12.06 & 70 & 0.50 & 3324 & 23.82 & 2591 & 18.57 & 45 & 0.32 & 4539 & 32.53 \\
\cline{2-20}
\multirow[t]{3}{*}{Olmo2-7B} & A & 512 & 3.67 & 711 & 5.10 & 1079 & 7.73 & 749 & 5.37 & 1679 & 12.03 & 1532 & 10.98 & 354 & 2.54 & 761 & 5.45 & 1945 & 13.94 \\
                                     & B & 9846 & 70.57 & 13065 & 93.64 & 9409 & 67.44 & 11072 & 79.36 & 12028 & 86.21 & 11569 & 82.92 & 10476 & 75.09 & 12993 & 93.13 & 9840 & 70.53 \\
                                     & Tie & 3594 & 25.76 & 176 & 1.26 & 3464 & 24.83 & 2131 & 15.27 & 245 & 1.76 & 851 & 6.10 & 3122 & 22.38 & 198 & 1.42 & 2167 & 15.53 \\
\cline{2-20}
\multirow[t]{3}{*}{Qwen2.5-7B} & A & 7016 & 50.29 & 11118 & 79.69 & 2902 & 20.80 & 6911 & 49.53 & 9676 & 69.35 & 6839 & 49.02 & 7544 & 54.07 & 9888 & 70.87 & 4586 & 32.87 \\
                                     & B & 4917 & 35.24 & 2094 & 15.01 & 8101 & 58.06 & 5826 & 41.76 & 4165 & 29.85 & 4568 & 32.74 & 5146 & 36.88 & 3552 & 25.46 & 8496 & 60.89 \\
                                     & Tie & 2019 & 14.47 & 740 & 5.30 & 2949 & 21.14 & 1215 & 8.71 & 111 & 0.80 & 2545 & 18.24 & 1262 & 9.05 & 512 & 3.67 & 870 & 6.24 \\
\cline{2-20}
\multirow[t]{3}{*}{Commandr-7B} & A & 3569 & 25.58 & 2430 & 17.42 & 3893 & 27.90 & 5300 & 37.99 & 4664 & 33.43 & 5005 & 35.87 & 4532 & 32.48 & 5280 & 37.84 & 5088 & 36.47 \\
                                     & B & 8868 & 63.56 & 8960 & 64.22 & 5789 & 41.49 & 7434 & 53.28 & 6869 & 49.23 & 5278 & 37.83 & 7987 & 57.25 & 6267 & 44.92 & 5697 & 40.83 \\
                                     & Tie & 1515 & 10.86 & 2562 & 18.36 & 4270 & 30.60 & 1218 & 8.73 & 2419 & 17.34 & 3669 & 26.30 & 1433 & 10.27 & 2405 & 17.24 & 3167 & 22.70 \\
\midrule
\multirow[t]{3}{*}{Mixtral-8x7B} & A & 1681 & 12.05 & 8348 & 59.83 & 600 & 4.30 & 1151 & 8.25 & 13239 & 94.89 & 393 & 2.82 & 1015 & 7.27 & 2529 & 18.13 & 1089 & 7.81 \\
                                     & B & 7577 & 54.31 & 4937 & 35.39 & 8618 & 61.77 & 10509 & 75.32 & 692 & 4.96 & 5322 & 38.15 & 8221 & 58.92 & 6204 & 44.47 & 8563 & 61.37 \\
                                     & Tie & 4694 & 33.64 & 667 & 4.78 & 4734 & 33.93 & 2292 & 16.43 & 21 & 0.15 & 8237 & 59.04 & 4716 & 33.80 & 5219 & 37.41 & 4300 & 30.82 \\
\cline{2-20}
\multirow[t]{3}{*}{Mistral-22B} & A & 319 & 2.29 & 3930 & 28.17 & 287 & 2.06 & 468 & 3.35 & 3456 & 24.77 & 1085 & 7.78 & 306 & 2.19 & 4568 & 32.74 & 474 & 3.40 \\
                                     & B & 12492 & 89.54 & 8299 & 59.48 & 12454 & 89.26 & 12782 & 91.61 & 9939 & 71.24 & 10818 & 77.54 & 12428 & 89.08 & 7969 & 57.12 & 12379 & 88.73 \\
                                     & Tie & 1141 & 8.18 & 1723 & 12.35 & 1211 & 8.68 & 702 & 5.03 & 557 & 3.99 & 2049 & 14.69 & 1218 & 8.73 & 1415 & 10.14 & 1099 & 7.88 \\
\cline{2-20}
\multirow[t]{3}{*}{Olmo2-32B} & A & 5725 & 41.03 & 5244 & 37.59 & 6709 & 48.09 & 6542 & 46.89 & 6602 & 47.32 & 5961 & 42.73 & 5694 & 40.81 & 6841 & 49.03 & 5983 & 42.88 \\
                                     & B & 5078 & 36.40 & 5578 & 39.98 & 5209 & 37.34 & 4782 & 34.27 & 6009 & 43.07 & 6316 & 45.27 & 3750 & 26.88 & 5626 & 40.32 & 4913 & 35.21 \\
                                     & Tie & 3149 & 22.57 & 3130 & 22.43 & 2034 & 14.58 & 2628 & 18.84 & 1341 & 9.61 & 1675 & 12.01 & 4508 & 32.31 & 1485 & 10.64 & 3056 & 21.90 \\
\cline{2-20}
\multirow[t]{3}{*}{Mixtral-8x22B} & A & 2019 & 14.47 & 3965 & 28.42 & 3768 & 27.01 & 3020 & 21.65 & 9143 & 65.53 & 2547 & 18.26 & 2619 & 18.77 & 6700 & 48.02 & 3266 & 23.41 \\
                                     & B & 4859 & 34.83 & 8357 & 59.90 & 8029 & 57.55 & 7172 & 51.40 & 3009 & 21.57 & 7448 & 53.38 & 7692 & 55.13 & 5657 & 40.55 & 8429 & 60.41 \\
                                     & Tie & 7074 & 50.70 & 1630 & 11.68 & 2155 & 15.45 & 3760 & 26.95 & 1800 & 12.90 & 3957 & 28.36 & 3641 & 26.10 & 1595 & 11.43 & 2257 & 16.18 \\
\midrule
\multirow[t]{3}{*}{Llama-70B} & A & 8434 & 60.45 & 9033 & 64.74 & 9001 & 64.51 & 7628 & 54.67 & 7282 & 52.19 & 8643 & 61.95 & 9171 & 65.73 & 8555 & 61.32 & 6698 & 48.01 \\
                                     & B & 5175 & 37.09 & 4384 & 31.42 & 4808 & 34.46 & 6187 & 44.34 & 6590 & 47.23 & 5188 & 37.18 & 4390 & 31.47 & 5198 & 37.26 & 6664 & 47.76 \\
                                     & Tie & 343 & 2.46 & 535 & 3.83 & 143 & 1.02 & 137 & 0.98 & 80 & 0.57 & 121 & 0.87 & 391 & 2.80 & 199 & 1.43 & 590 & 4.23 \\
\cline{2-20}
\multirow[t]{3}{*}{Qwen2.5-72B} & A & 7678 & 55.03 & 7368 & 52.81 & 7331 & 52.54 & 8250 & 59.13 & 8105 & 58.09 & 6544 & 46.90 & 7865 & 56.37 & 8221 & 58.92 & 6525 & 46.77 \\
                                     & B & 4609 & 33.03 & 3773 & 27.04 & 5598 & 40.12 & 5124 & 36.73 & 4011 & 28.75 & 5375 & 38.52 & 4646 & 33.30 & 4344 & 31.14 & 5128 & 36.75 \\
                                     & Tie & 1665 & 11.93 & 2811 & 20.15 & 1023 & 7.33 & 578 & 4.14 & 1836 & 13.16 & 2033 & 14.57 & 1441 & 10.33 & 1387 & 9.94 & 2299 & 16.48 \\
\cline{2-20}
\multirow[t]{3}{*}{Commandr-104B} & A & 10146 & 72.72 & 8455 & 60.60 & 11761 & 84.30 & 10095 & 72.36 & 9768 & 70.01 & 10976 & 78.67 & 11133 & 79.80 & 10760 & 77.12 & 10011 & 71.75 \\
                                     & B & 2072 & 14.85 & 1594 & 11.42 & 1357 & 9.73 & 1909 & 13.68 & 2547 & 18.26 & 2001 & 14.34 & 1342 & 9.62 & 2176 & 15.60 & 1361 & 9.75 \\
                                     & Tie & 1734 & 12.43 & 3903 & 27.97 & 834 & 5.98 & 1948 & 13.96 & 1637 & 11.73 & 975 & 6.99 & 1477 & 10.59 & 1016 & 7.28 & 2580 & 18.49 \\
\bottomrule
\end{tabular}}
\caption{Distribution of LLM Comparisons across each quality dimension and prompt configuration.}
\label{tab:human-llm-comp-dist}
\end{table*}

The dataset used in our experiments, Webis-ArgQuality-20, was created by~\citet{gienapp-etal-2020-efficient} and is available under a Creative Commons Attribution 4.0 International (CC BY 4.0) license. It is publicly accessible at \url{https://zenodo.org/records/3780049}.

The models employed in our experiments, as listed in Appendix \ref{sec:appendix-llms-version}, were used in accordance with their respective licenses and applicable usage policies. The code used in our experiments, together with the LLM-generated pairwise judgments and the Bradley-Terry rankings derived from them, is released under a Creative Commons Attribution 4.0 International (CC BY 4.0) license. Redistribution of model-generated outputs is believed to be permitted under the applicable licenses of the models used; users remain responsible for complying with the original model licenses and usage policies.

\section{Distribution of Human and LLM Comparisons}
\label{sec:dist-humans-llms}

\bocr{The Webis-ArgQuality-20 corpus contains 13952 comparisons. For the logic dimension, 6222 are predicted as A (44.60\%), 5719 as B (40.99\%), and 2011 as Tie (14.41\%); for rhetoric, 5833 are A (41.81\%), 5660 are B (40.57\%), and 2459 are Tie (17.62\%); and for dialectic, 6005 are A (43.04\%), 5767 are B (41.33\%), and 2180 are Tie (15.62\%). This means that, across the comparisons in the dataset, humans choose A and B in a more or less comparable and balanced quantity of times, while Ties occur less frequently.}

\bocr{Table~\ref{tab:human-llm-comp-dist} reports the LLM comparisons across each quality dimension and prompting method, reporting the majority-vote distribution across the three runs described in Section~\ref{s:experimental_setup} for each LLM configuration~\footnote{Given the amount of configurations (12 LLMs, 3 prompts, and 3 quality dimensions) We report the distribution (\texttt{distplots}) of the LLM-derived BT scores in the repository of the paper.}. Note that some models, such as Mistral-7B for small sizes and Mixtral-8x7B and Mixtral-8x22B for medium sizes, predict a higher quantity of Ties under zero-shot and CoT prompting, reaching up to 92.06\% for Mistral-7B with CoT prompting. When using few-shot prompting, this tendency is reversed; however, the model tends to favor one argument more strongly. For example, in the logic dimension, B is usually chosen in 94.64\% of the cases, while in rhetoric, A is chosen in 85.30\% of the cases. Tie is also chosen more often under few-shot prompting, with 62.97\% of the cases, although this is lower than the Tie rates observed with zero-shot and CoT prompting, which are 69.75\% and 74\%, respectively, for these dimensions. Larger models predict fewer Ties across all prompts and dimensions while predicting A more often than B.}

\end{document}